\documentclass[journal]{IEEEtai}

\usepackage[colorlinks,urlcolor=blue,linkcolor=blue,citecolor=blue]{hyperref}
\usepackage{color,array}
\usepackage{graphicx}
\usepackage{amsmath,amsfonts,amssymb,amsthm}
\usepackage{algorithm}
\usepackage[noend]{algpseudocode}
\usepackage{booktabs}
\usepackage{multirow}
\usepackage{microtype}

\begin{document}

\title{Bridging the Semantic-Utility Gap in Multimodal RAG via Generator-in-the-Loop Alignment}

\author{Zhan-Lun Chang, \IEEEmembership{Student Member, IEEE},
Dong-Jun Han, \IEEEmembership{Member, IEEE},
Seyyedali Hosseinalipour, \IEEEmembership{Senior Member, IEEE},
Mung Chiang, \IEEEmembership{Fellow, IEEE},
and Christopher G. Brinton, \IEEEmembership{Senior Member, IEEE}%
\thanks{Z.-L. Chang, M. Chiang, and C. G. Brinton are with the Elmore Family School of Electrical and Computer Engineering, Purdue University, West Lafayette, IN 47907, USA.}%
\thanks{D.-J. Han is with the Department of Computer Science and Engineering, Yonsei University, Seoul 03722, Republic of Korea.}%
\thanks{S. Hosseinalipour is with the Department of Electrical Engineering, University at Buffalo--SUNY, Buffalo, NY 14260, USA.}}

\markboth{IEEE Transactions on Artificial Intelligence}
{Multimodal RAG}

\makeatletter
\def\@IEEEpubidpullup{4\baselineskip}
\makeatother
\IEEEpubid{\parbox{0.95\textwidth}{\centering\scriptsize This work has been submitted to the IEEE for possible publication. Copyright may be transferred without notice, after which this version may no longer be accessible.}}

\maketitle

\begin{abstract}
Vision-language models (VLMs) augmented with retrieval-augmented generation (RAG) benefit from access to external evidence. However, standard retrievers and rerankers optimize for semantic similarity rather than answer utility, creating a preference gap: documents that appear relevant may not help the generator produce a correct answer. Motivated by this, we propose a two-stage \emph{generator-in-the-loop alignment framework} that closes this gap without human document-level relevance annotations. Our framework consists of two stages: in Stage~1, a VLM generates a hypothetical text passage from the image-query pair, which is used as the retrieval query for dense text search, bridging the image-to-text modality gap. In Stage~2, a cross-encoder reranker adapted with low-rank adaptation (LoRA) is fine-tuned using answer-supervised preference pairs mined from the frozen VLM: given the dataset answer label, a candidate document is labeled positive if the VLM produces the correct answer when given that document as context, and negative otherwise. This generator-guided signal is compatible with multiple alignment loss functions, including contrastive (triplet) loss, pairwise direct preference optimization (DPO), and supervised fine-tuning (SFT), and supports periodic re-mining to refresh preference pairs as the reranker improves. Experiments on VQA-X and A-OKVQA with Qwen3.5-2B and Qwen3-VL-4B-Instruct show that our proposed framework consistently outperforms rank-order, random, and REPLUG-style likelihood baselines under various alignment losses and pool size settings, suggesting that answer-level generator feedback is an effective supervision signal for preference alignment.
\end{abstract}

\begin{IEEEImpStatement}
    RAG-enabled multimodal systems are becoming a key component of knowledge-intensive visual reasoning applications, where retrieving misleading evidence can cause models to produce confident yet incorrect answers. This work addresses a fundamental limitation of existing multimodal RAG pipelines by shifting retrieval optimization from semantic similarity toward downstream answer utility. Rather than relying on costly human document-level relevance annotations, the proposed \textit{generator-in-the-loop alignment framework} leverages a frozen VLM to automatically identify evidence that improves answer correctness and uses this signal to train a lightweight LoRA-adapted reranker. The framework is compatible with multiple alignment objectives and consistently improves performance across different datasets and model architectures, demonstrating that answer-level supervision is a more effective retrieval objective than semantic similarity alone. By improving the reliability and faithfulness of multimodal RAG systems, this approach has the potential to benefit high-stakes applications such as medical decision support, scientific knowledge retrieval, education, document understanding, and other VLM-empowered systems that require trustworthy multimodal reasoning while reducing the dependence on expensive manual supervision.
\end{IEEEImpStatement}

\IEEEpubidadjcol

\section{Introduction}

\IEEEPARstart{L}{arge} language models (LLMs) and multimodal vision-language models (VLMs) have fundamentally transformed artificial intelligence, demonstrating unprecedented capabilities in natural language understanding, visual reasoning, and content generation \cite{yuan2025local, fang2026bridging}. However, despite their scale, these models remain plagued by intrinsic limitations: they are prone to hallucinations, restricted by static parametric knowledge cutoffs, and lack transparency in their reasoning processes \cite{huang2025survey}. 

To mitigate these issues, retrieval-augmented generation (RAG) has emerged as the de facto standard by replacing reliance on static parametric knowledge alone with the retrieval of external evidence at inference time \cite{lewis2020retrieval,borgeaud2022improving}. By grounding generation in retrieved evidence, RAG systems significantly enhance factual accuracy and interpretability, particularly in knowledge-intensive domains where hallucination remains a persistent reliability concern \cite{mcintosh2023culturally}.

The design of RAG systems has evolved from simple retrieve-then-generate pipelines to architectures that incorporate query rewriting, iterative retrieval, retrieval verification, and multi-step reasoning \cite{gao2023retrieval, khattab2023dspy, ma2023query}. While these techniques improve candidate retrieval and evidence refinement, retrievers and rerankers continue to optimize semantic similarity rather than the generator's downstream objective. Consequently, a document that appears highly relevant to the query may still fail to provide the evidence required for the generator to produce the correct answer.

More critically, semantically relevant documents can contain incorrect or misleading information that steers the generator toward an erroneous answer. Such documents may therefore increase, rather than reduce, the likelihood of hallucination \cite{cuconasu2024power}. A reranker must thus distinguish documents that help the generator produce the correct answer from those that are semantically relevant but unhelpful or misleading. Training a reranker to make this distinction typically requires document-level relevance annotations; however, manually determining whether each candidate document improves the generator's answer correctness requires costly, task-specific annotation at scale. This motivates the central question of this work:

\begin{quote}
\textit{\textbf{Q:} How can we train a reranker to identify evidence that improves the generator's answer correctness using only ground-truth answer labels with no human document-level relevance annotations?}
\end{quote}

Answering this question becomes more challenging in \textit{multimodal RAG} (MM-RAG), where retrieval must account for information contained in both visual and textual inputs. Specifically, visual embeddings optimized for general image-text alignment may fail to preserve fine-grained visual information, such as object relationships or structural details in scientific diagrams, that is necessary to retrieve evidence for downstream reasoning. This mismatch between the multimodal query and the retrieved textual evidence compounds the misalignment between retrieval relevance and answer correctness. Prior approaches, such as ATLAS \cite{izacard2023atlas} and REPLUG \cite{shi2024replug}, attempt to align retrieval relevance with the generator's downstream objective through likelihood- or perplexity-based training signals. However, these signals measure how a retrieved document changes the probability of a target output rather than whether the generator actually produces the correct answer. Therefore, MM-RAG requires addressing two coupled problems: (i) translating multimodal inputs into effective retrieval queries and (ii) selecting retrieved evidence based on its contribution to downstream answer correctness.

To jointly address these problems, we propose a two-stage \textit{generator-in-the-loop alignment framework}. In Stage~1, a frozen VLM converts the image-question pair into a hypothetical textual rationale using hypothetical document embeddings (HyDE). This rationale is concatenated with the original question and encoded by a \textit{Sentence Transformer} \cite{zhang2025qwen3} to retrieve a broad pool of candidate documents using Facebook AI similarity search (FAISS). Stage~2 addresses what we term the \textit{Preference Gap}: conventional rerankers prioritize documents based on semantic relevance, whereas the generator requires documents that help it produce the correct answer. A semantically relevant document may therefore lack the evidence needed to answer the question \cite{wang2024learning} or may even act as a distractor that degrades performance below the closed-book baseline \cite{liu2024lost, yoran2024making}. To directly align reranking with answer correctness, we provide each candidate document to the frozen VLM as context and compare the generated answer with the ground-truth answer. A document is labeled \textit{positive} if the model produces the correct answer and \textit{negative} otherwise, yielding preference pairs without human document-level relevance annotations. We use these pairs to fine-tune a cross-encoder reranker through low-rank adaptation (LoRA) and evaluate the same generator-derived supervision under contrastive learning, direct preference optimization (DPO), and supervised fine-tuning (SFT). By comparing the same supervision signal across three training objectives, we evaluate whether reranking based on the generator's answer correctness consistently improves evidence selection independent of the specific training objective.

The main contributions of this paper can be summarized as follows:
\begin{enumerate}
    \item We propose a two-stage multimodal RAG framework that separates multimodal query translation from generator-aligned reranking. In Stage~1, a frozen VLM generates a hypothetical textual rationale from the image-question pair using HyDE. The rationale is concatenated with the question and encoded by a Sentence Transformer to retrieve a broad pool of candidate documents using FAISS.
    
    \item We introduce an \textit{answer-supervised preference mining} procedure that trains the Stage~2 reranker without human document-level relevance annotations. Specifically, for each candidate document, the frozen VLM generates an answer using the document as context; documents that lead to correct and incorrect answers are used to construct positive-negative preference pairs for reranker training.

    \item We use the generator-derived preference pairs to fine-tune a LoRA-adapted cross-encoder reranker under three training objectives: contrastive triplet loss, DPO, and SFT. Experiments on VQA-X and A-OKVQA with Qwen3.5-2B and Qwen3-VL-4B-Instruct show that generator-guided preference mining consistently outperforms rank-order, random, and REPLUG-style likelihood preference mining across the considered training objectives and various candidate pool sizes.

    \item We develop an iterative preference-alignment mechanism with \textit{periodic preference re-mining} that dynamically reconstructs generator-derived preference pairs as the reranker evolves, thereby keeping the training signal aligned with the reranker’s current ranking behavior rather than relying on a fixed set of initially mined pairs. Experiments across various generators and datasets demonstrate consistent improvements over fixed-pair training, with gains of up to 2.58 percentage points.

\end{enumerate}

\vspace{-2mm}
\section{Related Work}
We organize the related literature into the following categories: (i) evolution of RAG; (ii) multimodal retrieval and the semantic gap; (iii) retriever-generator alignment.

\vspace{-2mm}
\subsection{Evolution of RAG}
RAG has emerged as a standard method for mitigating the limitations of LLMs/VLMs, specifically their susceptibility to hallucinations and outdated parametric knowledge \cite{lewis2020retrieval, guu2020retrieval}. Following recent surveys \cite{gao2023retrieval, gupta2024comprehensive}, we summarize the progression of RAG through three phases: \textit{Naive RAG methods}, \textit{Advanced RAG methods}, and \textit{Modular RAG methods}.

The initial phase, \textit{Naive RAG methods}, followed a linear ``retrieve-then-generate'' protocol. These systems utilized sparse (BM25) or dense passage retrieval (DPR) retrievers to fetch context, which was then directly concatenated with the user query \cite{karpukhin2020dense}. While effective for simple factual retrieval, Naive RAG methods struggle with low precision and noise induction, where irrelevant retrieved context degrades the generator's performance \cite{shuster2021retrieval}. To mitigate these shortcomings, approaches such as \textit{Chain-of-Note} \cite{yu2024chain} were introduced, teaching models to explicitly generate \textit{notes} assessing the relevance of retrieved documents before answering, thereby improving robustness against noisy inputs.

To further improve performance, \textit{Advanced RAG} methods introduced sophisticated pre-retrieval and post-retrieval optimizations. Specifically, pre-retrieval techniques, such as query rewriting \cite{ma2023query}, multi-aspect query retrieval \cite{xu2026maq}, and HyDE \cite{gao2023precise}, aim to align the user's intent with the document space. HyDE in particular prompts a language model to generate a hypothetical answer-oriented document from the query and uses that generated text, rather than the raw query, as the retrieval input; because this text resembles the corpus documents more closely, its embedding can have higher similarity to answer-bearing passages and improve recall without retrieval-specific training. Also, post-retrieval strategies focused on refining the context window. Notably, reranking models \cite{nogueira2019passage} and cross-encoder filtering \cite{sun2023chatgpt} were developed to identify the most relevant retrieved passages, thereby reducing irrelevant context before generation. Despite these improvements, recent studies on the ``lost in the middle'' phenomenon \cite{liu2024lost} suggest that simply stuffing context windows is insufficient, as models struggle to access information buried in the middle of long retrieval sequences.

Most recently, the field has shifted toward \textit{modular and active RAG} \cite{oche2025systematic}. In this paradigm, the pipeline is decomposed into independent, swappable modules orchestrated by programming frameworks such as DSPy \cite{khattab2023dspy}, which optimizes the flow of information between modules. Beyond static retrieval, recent frameworks further employ \textit{active} strategies. For instance, FLARE \cite{jiang2023active} drafts upcoming generated text and triggers retrieval when low-confidence tokens suggest that additional evidence is needed, while CRAG (corrective RAG) \cite{yan2024corrective} introduces a lightweight retrieval evaluator that triggers a web search fallback when the retrieved documents are deemed incorrect. These systems improve \textit{when} and \textit{how} retrieval is invoked, but evidence selection often remains based on \textit{semantic similarity}, meaning how closely a retrieved document matches the query in topic, wording, or embedding space. However, a semantically similar document may still be unhelpful or even misleading for answering the question. Therefore, evidence should instead be selected according to its \textit{downstream answer utility}, namely whether providing that document as context actually helps the generator produce the correct answer.

\subsection{Multimodal Retrieval and the Semantic Gap}
The success of RAG in the textual domain has extended to large multimodal models, giving rise to MM-RAG \cite{zhu2024minigpt, liu2023visual}, systems that retrieve external textual or visual evidence to support reasoning over image–text queries. Unlike text-only RAG, MM-RAG must align heterogeneous query inputs, such as images and natural-language questions, with evidence stored in an external knowledge base. Standard retrieval architectures commonly rely on dual-encoder backbones such as CLIP \cite{radford2021learning} to map visual and textual content into a shared semantic space. Meanwhile, foundational models such as Flamingo \cite{alayrac2022flamingo} demonstrated the effectiveness of conditioning generation on interleaved image–text inputs. However, neither multimodal conditioning nor coarse image–text alignment alone guarantees that the retrieved evidence contains the fine-grained information needed for downstream reasoning. This leads to a critical \textit{semantic gap}: visual embeddings optimized for general image–caption matching may overlook subtle objects, relationships, and structural details that are essential for retrieving answer-supporting evidence \cite{oche2025systematic}. For example, Qwen-VL \cite{bai2023qwen} attempts to preserve finer visual information through resolution-aware visual encoders, but still relies on conventional retrieval mechanisms. Recent efforts have therefore sought to narrow this gap by integrating stronger generative or fine-grained visual representations into retrieval. RA-CM3 \cite{yasunaga2022retrieval} was among the first to jointly retrieve and generate mixed-modal content within a unified architecture. More recently, ColPali \cite{faysse2025colpali} avoids the lossy conversion of document pages into text by using a VLM to produce multi-vector embeddings for page patches, enabling late-interaction retrieval that better preserves visual layout and textual semantics.

Our work takes a complementary approach to these methods by decoupling multimodal query alignment from evidence-utility estimation. Rather than requiring the retriever itself to learn a shared representation that preserves all task-relevant visual and textual information, we use a frozen VLM to translate the image–question pair into a hypothetical textual rationale. This converts the multimodal query into the same modality as the document corpus, enabling effective text-based dense retrieval without training a dedicated multimodal retriever. However, resolving the modality gap alone does not ensure that the retrieved documents will help the generator answer correctly. We therefore introduce a second stage in which a separately trained cross-encoder reranker is aligned with the frozen generator’s downstream behavior, prioritizing documents according to their answer utility rather than semantic similarity.

\vspace{-3mm}
\subsection{Retriever-Generator Alignment}
The most significant challenge in RAG is the \textit{alignment gap} between retrieval or reranking objectives and the generator's reasoning objective \cite{gao2023retrieval}. In particular, retrievers are typically optimized for query-document semantic similarity, an objective decoupled from whether the retrieved context actually helps the generator produce the correct answer. This mismatch frequently leads to failure in complex reasoning tasks \cite{he2024retrieving}, despite evidence that LLMs can effectively utilize in-context knowledge when the right documents are retrieved \cite{ram2023context}.

Several methods have sought to reduce the alignment gap by incorporating generator feedback into retrieval training. For instance, Atlas \cite{izacard2023atlas} jointly trains the retriever and generator using generation perplexity, while REPLUG \cite{shi2024replug} treats the language model as a black box and trains the retriever using each document’s effect on the probability of the ground-truth continuation. These methods make retrieval more generator-aware, but their likelihood-based signals remain indirect proxies for downstream correctness and were developed primarily for text-only generation settings.

BGM \cite{ke2024bridging} addresses a related \textit{preference gap}, in which the retriever’s highest-ranked documents do not necessarily produce the best generation outcomes. Rather than updating the frozen retriever and generator, BGM inserts a trainable sequence-to-sequence bridge between them. The bridge is trained using supervised learning on synthesized silver passages and reinforcement learning to rerank and reorganize the retrieved context. Although this approach moves closer to task-level optimization, it introduces an additional large trainable component into the RAG pipeline.

Our framework instead aligns reranking directly with the deployed generator’s answer behavior: for each candidate document, the frozen VLM is prompted to produce the constrained dataset answer output, either yes/no or one of the provided options, and the document is labeled useful when that output matches the dataset-provided ground-truth. The resulting signal directly measures whether a document helps the generator answer correctly, rather than whether it increases the likelihood of the ground-truth answer when preceding answer tokens are supplied. These generator-specific document-level preferences are then distilled into a lightweight cross-encoder reranker. Our method therefore complements conventional semantic rerankers \cite{nogueira2019passage, sun2023chatgpt}, likelihood- and reward-based retrieval alignment methods \cite{izacard2023atlas, shi2024replug, ke2024bridging}, and multimodal retrieval approaches that primarily address cross-modal representation alignment \cite{yasunaga2022retrieval, faysse2024colpali}.

Our proposed supervision mechanism is related to \textit{LLM-as-judge} and synthetic-data methods. LLM-as-judge approaches use language models to assign general quality scores or preferences to open-ended outputs \cite{liu2023g}, while self-generated instruction methods synthesize training examples to reduce dependence on manual annotation \cite{wang2023self}. In our setting, however, the VLM neither produces a general evaluation nor generates new instructions. Instead, it reveals whether a particular retrieved document causes the same frozen generator used at deployment to produce the correct answer. The resulting labels differ from general-purpose LLM judgments or synthetic training data: they provide generator-specific, document-level utility signals tailored explicitly to reranker alignment.

\begin{figure*}[t]
    \centering
    \includegraphics[width=\linewidth]{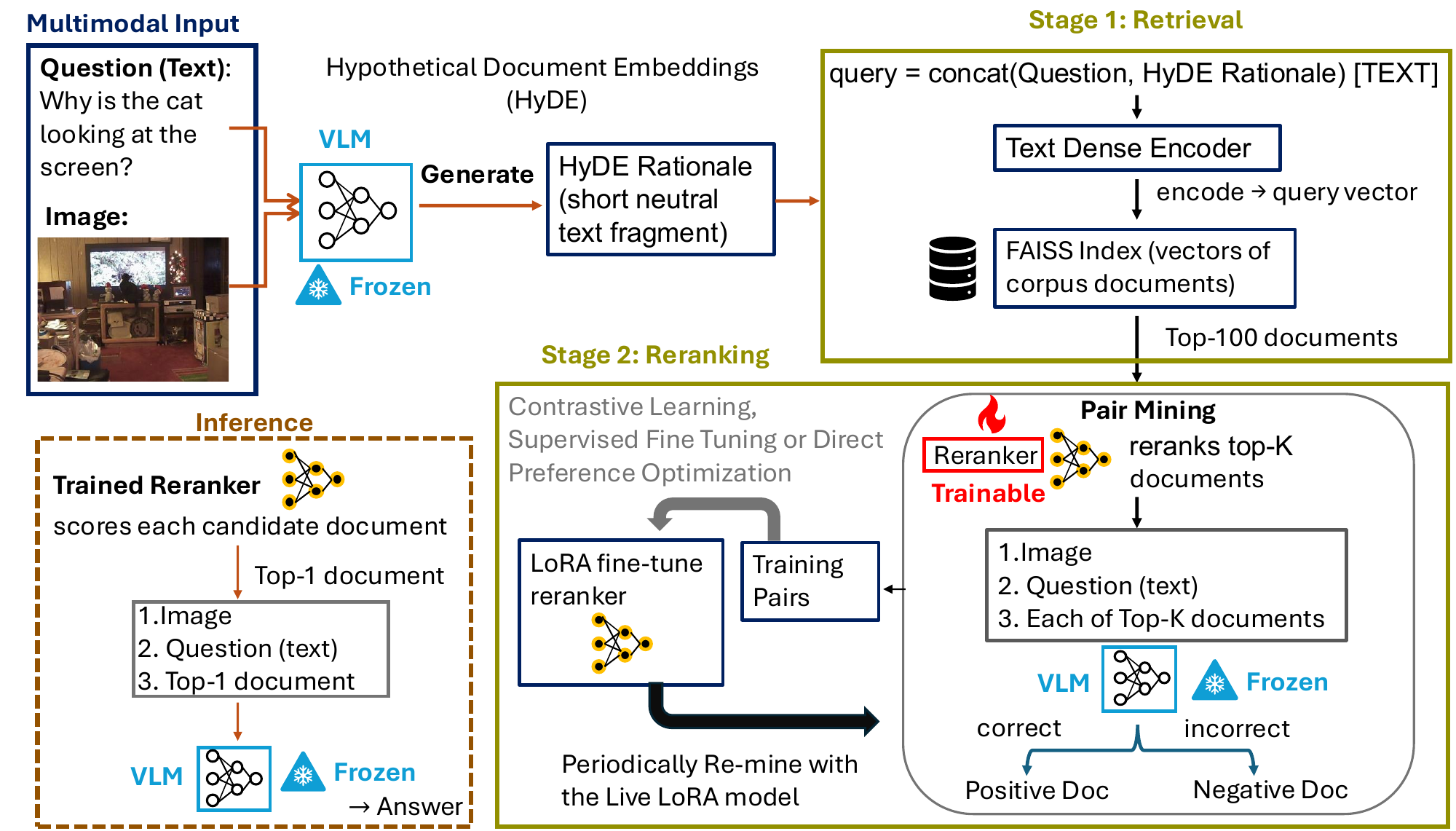}
    \vspace{-7mm}
    \caption{A schematic of our proposed two-stage multimodal RAG system. \textbf{Stage~1 (Retrieval):} given a multimodal input $(v, q)$, the frozen VLM $\mathcal{G}$ generates a short HyDE rationale $h$; the composite query $q_\mathrm{text} = q \oplus h$ is encoded by a text dense encoder and used to retrieve the top-$k$ corpus documents from a FAISS index. \textbf{Stage~2 (Reranking):} a LoRA-adapted cross-encoder scores each of the $k$ candidate documents and is fine-tuned on preference pairs $(d^+, d^-)$ mined by querying the frozen VLM for answer correctness. Preference pairs are periodically re-mined using the live LoRA model, and the reranker can be trained under contrastive (triplet), SFT, or DPO losses. \textbf{Inference:} the trained reranker selects the top-1 document, which is passed together with the image and question to the frozen VLM to produce the final answer.}
    \label{fig:system_architecture}
    \vspace{-3mm}
\end{figure*}

\vspace{-2mm}
\section{Proposed Architecture} \label{sec:proposed_architecture}

In this section, we present the architecture of our two-stage MM-RAG system and the alignment framework used to close the gap between candidate ranking and answer utility. The key idea is to separate broad multimodal recall from generator-specific utility alignment: Stage~1 uses the frozen deployment generator as an image-to-text bridge for dense retrieval, while Stage~2 uses the same frozen generator to mine answer-level preferences for reranker training. Table~\ref{tab:notation} summarizes the mathematical notations used throughout.

\begin{table}[t]
\caption{Mathematical Notations}
\label{tab:notation}
\vspace{-2mm}
\centering
\begin{tabular}{cl}
\toprule
Symbol & Meaning \\
\midrule
$v$ & Input image \\
$q$ & Natural-language question string \\
$a_\mathrm{gold}$ & Ground-truth answer \\
$\mathcal{G}$ & Frozen VLM generator \\
$h = \mathcal{G}(v,q)$ & HyDE hypothetical rationale text \\
$q_\mathrm{text} = q \oplus h$ & Concatenated query string \\
$\phi(\cdot)$ & Qwen3-Embedding-8B text encoder \\
$\mathbf{e}_q = \phi(q_\mathrm{text})$ & $\ell_2$-normalized query embedding \\
$\mathcal{I}$ & FAISS \texttt{IndexFlatIP} index \\
$\mathcal{C}_q$ & Stage~1 top-$K$ candidate pool \\
$K$ & Stage~1 retrieval depth (default 100) \\
$K'$ & Stage~2 preference-mining pool cap, with $K' \le K$  \\
$E_{\theta}$ & Cross-encoder ($\theta = \theta_\mathrm{base} \cup \theta_\mathrm{LoRA}$) \\
$E_\mathrm{ref}$ & Frozen reference reranker (no LoRA) \\
$s_\theta(q_\mathrm{text},d)$ & Scalar relevance logit for pair $(q_\mathrm{text},d)$ \\
$\mathcal{D}$ & Mined preference dataset \\
$d$ & A text document (e.g., textual justification or rationale) \\
$d^+$ & Positive (preferred) document \\
$d^-$ & Negative (rejected) document \\
$\mathcal{S}_+,\,\mathcal{S}_-$ & Correct / incorrect candidate sets \\
$B_m$ & VLM mini-batch size during mining \\
$N$ & Re-mining frequency (epochs between re-mines) \\
$\eta$ & AdamW learning rate \\
$\beta$ & DPO inverse-temperature \\
$m$ & Triplet margin \\
$\Psi$ & Persistent VLM prediction cache \\
\bottomrule
\end{tabular}
\vspace{-4mm}
\end{table}

\vspace{-2mm}
\subsection{Problem Formulation}

We formulate reranking as an answer-utility alignment problem rather than a standard relevance-ranking problem. Formally, we presume that the corpus $\mathcal{D}$ consists of text documents $d \in \mathcal{D}$ (e.g., textual justifications for VQA-X or rationales for A-OKVQA) that serve as retrieval targets: each document provides external textual evidence that the generator $\mathcal{G}$ can use as context to help answer a question. Each query consists of an image $v$ and a natural-language question $q$; the dataset also provides a ground-truth answer $a_\mathrm{gold}$ that is available only during preference mining. The textual query representation used by the retriever and reranker is denoted $q_\mathrm{text}$ and is formed by concatenating the question $q$ with the hypothetical rationale $h$ generated by $\mathcal{G}$. Retrieving with $q_\mathrm{text}$ yields the candidate pool $\mathcal{C}_q$ consisting of $K$ documents. Given a frozen generator $\mathcal{G}$, a query $(v, q)$, and a candidate document $d \in \mathcal{C}_q$, we define the \emph{answer utility} of $d$ as the binary indicator
\begin{equation}
    u(d) = \mathbf{1}\bigl[\mathcal{G}(v,\, q,\, d) = a_\mathrm{gold}\bigr],
    \label{eq:utility}
\end{equation}
which equals 1 if and only if $\mathcal{G}$ produces the correct answer when given $d$ as context. We further partition the candidate pool $\mathcal{C}_q$ accordingly into a set of \emph{useful} documents $\mathcal{S}_+ = \{d \in \mathcal{C}_q : u(d) = 1\}$ and a set of \emph{harmful} documents $\mathcal{S}_- = \{d \in \mathcal{C}_q : u(d) = 0\}$.

Over this pool, a pre-trained cross-encoder $E_\theta$ scores each pair $(q_\mathrm{text}, d)$ with a relevance logit $s_\theta(q_\mathrm{text}, d)$ based on \emph{semantic similarity}, selecting $d^* = \arg\max_{d \in \mathcal{C}_q}\, s_\theta(q_\mathrm{text}, d)$. However, semantic similarity does not imply answer utility: a document semantically close to the query may fail to support correct reasoning, while a document that appears less similar may provide exactly the missing evidence. This gives rise to the \textbf{preference gap}, which can be interpreted as follows:
\begin{equation}
    E_\theta \text{ is misaligned when } d^* \notin \mathcal{S}_+.
    \label{eq:gap}
\end{equation}

\textbf{Research Problem.} Given a candidate pool $\mathcal{C}_q$ and a frozen generator $\mathcal{G}$, we aim to learn reranker parameters $\theta$ such that
\begin{equation}
    s_\theta(q_\mathrm{text}, d^+) > s_\theta(q_\mathrm{text}, d^-)
    \quad \forall\, d^+ \in \mathcal{S}_+,\; d^- \in \mathcal{S}_-,
    \label{eq:objective}
\end{equation}
\emph{without} any human-annotated document relevance labels. In this paradigm, the supervision signal is the answer utility $u(d)$, which is computed by querying the frozen $\mathcal{G}$ and comparing its output to the dataset-provided answer label.

\vspace{-2mm}
\subsection{Architecture Overview}

Our proposed framework is built on a two-stage pipeline, illustrated in Fig.~\ref{fig:system_architecture}. This stage-wise decomposition is intentional: Stage~1 is optimized for recall across the text corpus, while Stage~2 learns which recalled evidence actually improves the frozen generator's answer behavior. More specifically, Stage~1 converts the multimodal input $(v, q)$ into a short hypothetical text rationale $h$ via HyDE, which is concatenated with the question to form the composite text query $q_\mathrm{text} = q \oplus h$. This query is encoded by a text encoder (e.g., Qwen3-Embedding-8B) $\phi$ and used to search a FAISS index $\mathcal{I}$, producing a candidate pool $\mathcal{C}_q$ of up to $K$ documents. Subsequently, Stage~2 takes $\mathcal{C}_q$ and applies a LoRA-adapted cross-encoder $E_\theta$ to rerank the candidates by assigning a scalar relevance logit $s_\theta(q_\mathrm{text},d)$ to each text pair $(q_\mathrm{text}, d)$. The reranker is fine-tuned on preference pairs mined by querying the frozen $\mathcal{G}$ for answer correctness; pairs are periodically re-mined using the live model to keep the training signal fresh. At inference, the top-scored document $d^* = \arg\max_{d \in \mathcal{C}_q} s_\theta(q_\mathrm{text},d)$ is passed together with $(v, q)$ to the frozen $\mathcal{G}$ for final answer generation.

One of our key design choices is to separate multimodal understanding from reranker training. Specifically, the frozen generator $\mathcal{G}$ is the only component that processes the image $v$: in Stage~1, it translates the image--question pair into a textual rationale that enables text-based retrieval, while in Stage~2, it evaluates whether each retrieved document helps produce the correct answer. In contrast, the cross-encoder $E_{\theta}$ operates exclusively on text, allowing it to leverage strong pre-trained language representations without requiring multimodal adaptation. This separation keeps the architecture lightweight: $\mathcal{G}$ remains frozen and provides both modality conversion and supervision, while only the text-based reranker is trained.

\vspace{-3mm}
\subsection{Datasets and Document Corpora}

\begin{figure}[t]
    \centering
    \includegraphics[width=\linewidth]{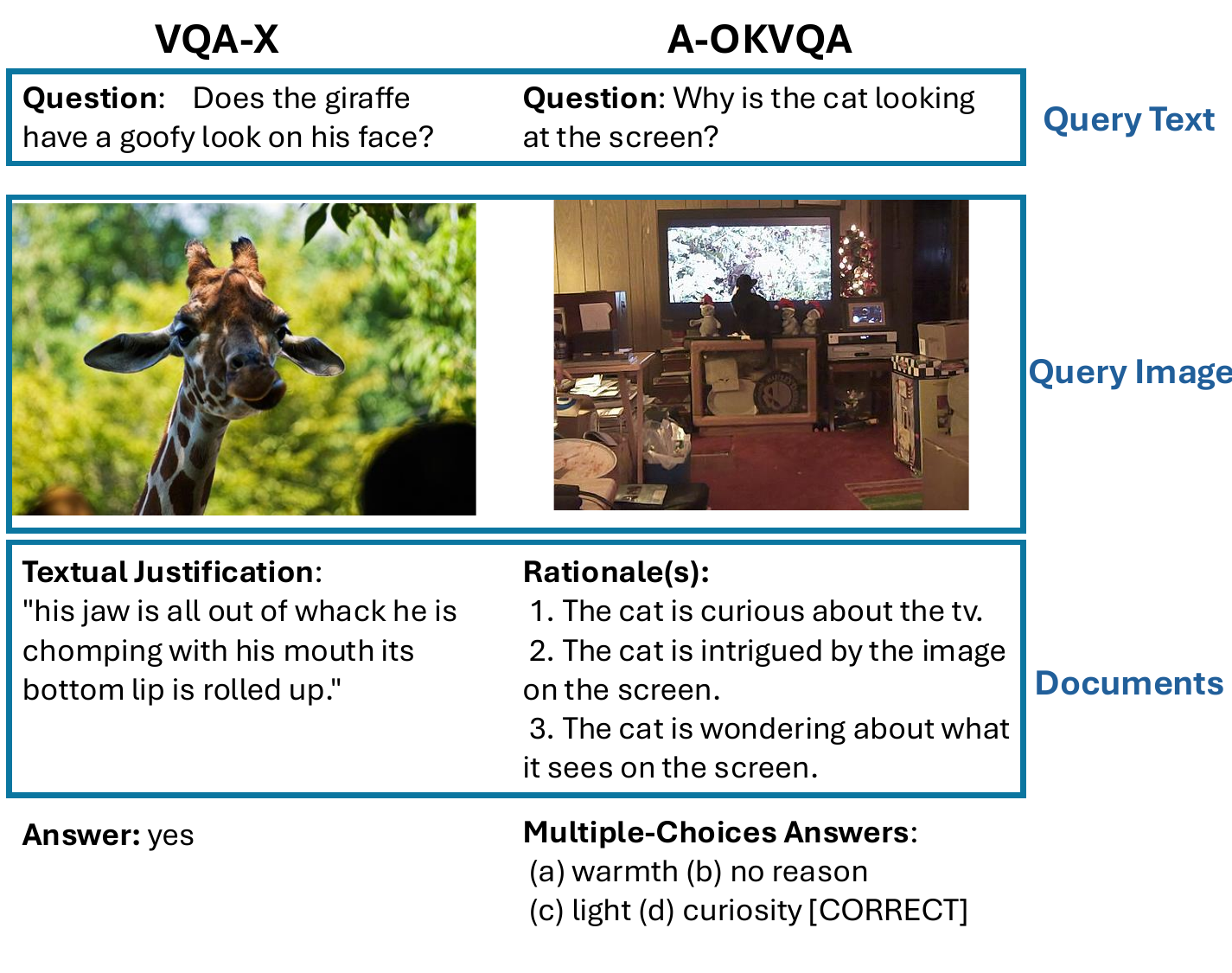}
     \vspace{-9mm}
    \caption{Dataset Integration into the MM-RAG Pipeline. 
    For each dataset (A-OKVQA and VQA-X), we map the original inputs to a unified query format comprising the image and question text. The external knowledge base is populated with the dataset-specific ground-truth explanations (rationales and justifications). The retrieval target is defined as the textual evidence required to bridge the reasoning gap between the visual query and the final answer.}
    \label{fig:dataset_pipeline}
    \vspace{-4mm}
\end{figure}

Fig.~\ref{fig:dataset_pipeline} illustrates what a real-world multimodal query looks like in the two datasets considered in this work: each query consists of an image $v$ and a natural-language question $q$, while the knowledge base is populated with dataset-specific textual documents (justifications for VQA-X, rationales for A-OKVQA) that serve as the retrieval targets for Stage~1. The following paragraphs briefly describe how each dataset defines the multimodal query, constructs the external knowledge base, and uses the retrieved evidence for answer generation.

\begin{itemize}
    \item \textbf{A-OKVQA (Augmented Outside Knowledge VQA):} This dataset focuses on visual question answering tasks that require commonsense or world knowledge beyond what is directly visible in the image. As shown in Fig.~\ref{fig:dataset_pipeline}, the pipeline accepts the query image (e.g., a cat watching TV) and the associated question text (e.g., Why is the cat looking at the screen?). The external database is populated with the dataset's \textit{Rationales}, which provide the necessary background information (e.g., The cat is intrigued by the image on the screen). The generator uses these retrieved rationales to select the correct option from multiple-choice answers or provide a direct response.

    \item \textbf{VQA-X (Visual Question Answering with eXplanations):} This dataset extends standard VQA by providing textual justifications for answers. As shown in Fig.~\ref{fig:dataset_pipeline}, the query is formed by the image (e.g., a giraffe) and the question (e.g., Does the giraffe have a goofy look on his face?). The \textit{Textual Justification} serves as the retrievable document. By retrieving these ground-truth reasons (e.g., ...his jaw is all out of whack...), the pipeline aligns the generator's output with the expected reasoning process.
\end{itemize}

To prevent test-set evidence leakage, the retrieval corpus is constructed exclusively from the training split and fixed before Stage~2 training. For A-OKVQA, the FAISS index contains the unique, non-empty rationales from the training set, whereas for VQA-X, it contains the unique, non-empty explanations associated with yes/no training examples. Stage~1 retrieves candidates from this fixed corpus and stores them in split-specific candidate files. Stage~2 then combines these files with the corresponding dataset split to recover the images, answer choices, and ground-truth labels required for preference mining and evaluation. In this way, the reranker is trained on candidates drawn only from the training-derived corpus, while its supervision is obtained from answer correctness rather than human document-level relevance annotations.

\vspace{-3mm}
\subsection{(Stage 1) HyDE-Based Modality Bridge and Dense Recall}

A fundamental challenge in multimodal RAG is the \textit{modality gap} between the multimodal query and the text-only retrieval space: the text encoder $\phi$ accepts only text, while the query is inherently multimodal, comprising an image $v$ and a natural-language question $q$. Using $q$ alone therefore discards potentially essential visual information from $v$, while training a dedicated cross-modal retriever would require substantial amounts of aligned image–text data and additional model adaptation.

To bridge this gap, we use HyDE~\cite{gao2023precise} as a multimodal-to-text conversion mechanism rather than merely as a generic query-expansion technique. Specifically, rather than searching with the raw question $q$ or training a separate multimodal retriever, the frozen generator $\mathcal{G}$ is first prompted with the full multimodal input $(v, q)$ to produce a \emph{hypothetical rationale}
\begin{equation}
    h = \mathcal{G}(v, q),
    \label{eq:hyde}
\end{equation}
where $h$ is a short, image-grounded passage that captures the visual information relevant to the question and resembles the type of evidence contained in the retrieval corpus. The rationale is then concatenated with the original question to form the composite textual query $q_\mathrm{text} = q \oplus h$. Since $q_{\text{text}}$ is entirely textual, it can be processed directly by $\phi$ without any multimodal adaptation. HyDE thus serves as an image-to-text modality bridge, projecting the task-relevant information in $v$ into the textual domain and enabling text-only dense retrieval for an inherently multimodal query.

The resulting query embedding is computed as $\mathbf{e}_q = \phi(q_\mathrm{text}) \in \mathbb{R}^n$, where $\phi$ prepends an instruction prefix and applies $\ell_2$ normalization. The corpus documents are encoded offline using the same encoder, and their embeddings are stored in a FAISS \texttt{IndexFlatIP} index $\mathcal{I}$ that supports exact inner-product search. Because each rationale or justification is already a short, self-contained unit, no document chunking is applied; instead, each document is indexed in its entirety. At query time, the top-$K$ candidates are retrieved as
\begin{equation}
    \mathcal{C}_q = \mathrm{top}\text{-}K\!\left(\mathbf{e}_q,\, \mathcal{I}\right).
    \label{eq:faiss_search}
\end{equation}
To avoid repeated VLM inference, the hypothetical rationale $h$ is cached per question identifier with atomic disk writes. The resulting candidate set is represented as $\mathcal{C}_q = \{(d_1, r_1), \ldots, (d_K, r_K)\}$, where $d_i$ denotes the $i$-th retrieved document and $r_i$ its retrieval rank. These candidate sets are then stored and passed to Stage~2 for generator-aligned reranking.

\begin{figure*}[t]
    \centering
    \includegraphics[width=0.9\linewidth]{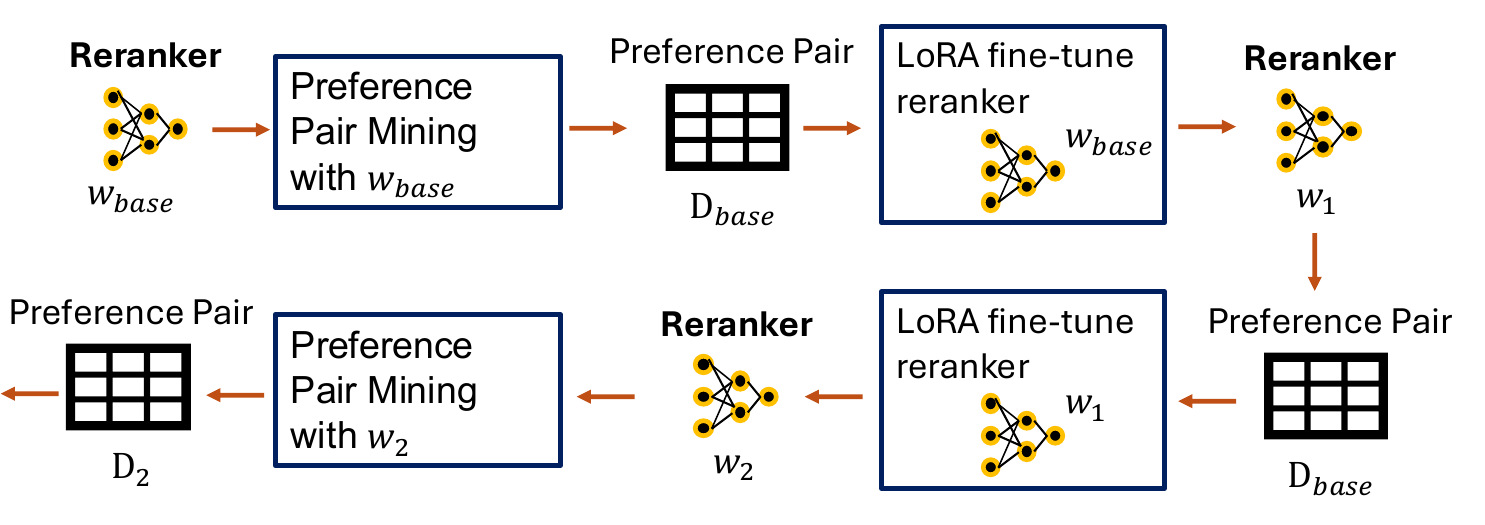}
     \vspace{-7mm}
    \caption{Iterative preference alignment with periodic re-mining. Training begins with the base reranker weights $w_\mathrm{base}$. Preference pairs are first mined under $w_\mathrm{base}$, producing dataset $\mathcal{D}_\mathrm{base}$. The reranker is then LoRA fine-tuned on $\mathcal{D}_\mathrm{base}$, advancing the weights from $w_\mathrm{base}$ to $w_1$ and then to $w_2$. After a fixed number of epochs, the preference pairs are re-mined under the current model $w_2$, yielding a refreshed dataset $\mathcal{D}_2$ that reflects the model's updated ranking. This cycle can keep the training signal aligned with the reranker's current behavior, although the benefit of additional refreshes may saturate once the model has stabilized.}
    \label{fig:lora_alignment}
\end{figure*}

\begin{figure*}[t]
    \centering
    \includegraphics[width=\linewidth]{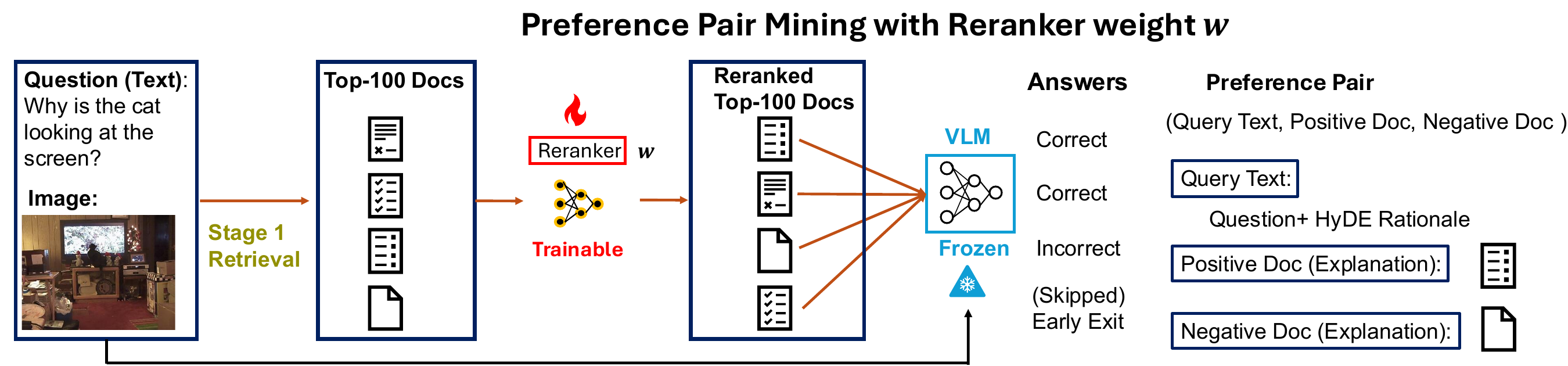}
     \vspace{-7mm}
    \caption{Generator-guided preference pair mining (proposed). Given a multimodal query $(v, q)$, Stage~1 retrieves 100 candidate documents. The trainable reranker with current weights $w$ re-scores those candidates, and Stage~2 retains the top-$K'=10$ subset for mining. The frozen VLM then evaluates each retained document in descending score order: if the generated answer is correct, the document is taken as $d^+$ and subsequent correct answers are skipped; if the answer is incorrect, the document is taken as $d^-$. Evaluation exits immediately once both $d^+$ and $d^-$ are found, avoiding unnecessary VLM calls. The resulting training triple is (query text $=$ question $\oplus$ HyDE rationale,\ $d^+$,\ $d^-$).}
    \label{fig:preference_mining}
    \vspace{-3mm}
\end{figure*}

\vspace{-3mm}
\subsection{(Stage 2) Generator-Guided Cross-Encoder Alignment}

Stage~2 is responsible for reranking the candidate pool $\mathcal{C}_q$ retrieved by Stage~1. Its key objective is to replace indirect supervision signals, such as semantic relevance or ground-truth-token likelihood, with direct feedback on whether the generator's constrained answer output matches the ground-truth. Specifically, the reranker is trained to prefer documents that enable the frozen generator $\mathcal{G}$ to produce the dataset-provided ground-truth answer $a_\mathrm{gold}$. To learn this preference while keeping training lightweight, we adapt a GTE-ModernBERT-base cross-encoder~\cite{warner2025smarter} with LoRA ($r = 16$, $\alpha = 32$, dropout $= 0.1$), where only the low-rank adapter weights $\theta_\mathrm{LoRA}$ are updated during training while the pre-trained encoder $\theta_\mathrm{base}$ is kept frozen. For each query–document pair, the cross-encoder jointly processes the composite textual query $q_{\text{text}}$ and the candidate document $d$, truncated to a maximum of 512 tokens, and produces a scalar score
\begin{equation}
    s_\theta(q_\mathrm{text}, d) = E_\theta\!\bigl([q_\mathrm{text};\, d]\bigr) \in \mathbb{R},
    \label{eq:score}
\end{equation}
This score represents the reranker’s learned estimate of how useful the document is for supporting the generator’s answer. To control VLM inference cost during preference mining, Stage~2 evaluates only the top-$K'$ prefix of the Stage~1 pool, where $K' \le K$.

Training supervision is derived automatically through \emph{preference mining}: for each training query, the current reranker $E_\theta$ first scores the Stage~1 candidates and retains only the top-$K'$ prefix for mining. A preference pair $(d^+, d^-)$ is then constructed, where $d^+$ denotes a preferred document and $d^-$ denotes a rejected one. All four mining strategies follow this same pairwise formulation but differ in how they assign the positive and negative labels. The proposed \textbf{generator-guided} strategy feeds each retained candidate document to the frozen VLM $\mathcal{G}$ together with the image $v$ and question $q$, and checks whether $\mathcal{G}$ produces the correct answer under the dataset answer label. The first retained candidate (in reranker-score order) that leads to a correct answer becomes $d^+$, and the first retained candidate that leads to an incorrect answer becomes $d^-$. If the retained pool does not contain both outcomes, the query is skipped and contributes no training pair for that epoch. The controlled baselines construct the same pairwise labels from different signals. The \textbf{REPLUG-style likelihood} baseline~\cite{shi2024replug} does not rely on constrained answer generation; instead, it scores each retained candidate according to the likelihood of $a_\mathrm{gold}$ when preceding ground-truth answer tokens are supplied and selects the highest- and lowest-scoring documents as $d^+$ and $d^-$. The \textbf{first} baseline requires no VLM: it uses only the current reranker $E_\theta$, assigning its highest-ranked candidate to $d^+$ and its lowest-ranked candidate to $d^-$. The \textbf{random} baseline ignores both reranker scores and VLM feedback: $d^+$ and $d^-$ are drawn uniformly at random from the retained pool. The mined pairs are used to fine-tune $E_\theta$ under one of three alignment loss functions: triplet margin loss, SFT, or pairwise DPO~\cite{rafailov2023direct}. To keep preference data fresh as the model improves, pairs are periodically re-mined against the current $E_\theta$ checkpoint rather than a fixed initial ranking. Section~\ref{sec:proposed_algorithm} provides pseudocode and walk-throughs for the training loop and mining strategies.

\vspace{-3mm}
\section{Reranker Alignment and Preference Mining}
\label{sec:proposed_algorithm}

Section \ref{sec:proposed_architecture} introduced the two-stage MM-RAG architecture, in which Stage 1 translates the multimodal query into text and retrieves a candidate document pool, while Stage 2 reranks these candidates according to their utility to the frozen generator. Building on this architecture, this section describes how the reranker is trained and evaluated. Specifically, we present the iterative LoRA-based alignment procedure with periodic re-mining, define the three alignment loss functions, and describe both the proposed generator-guided preference-mining strategy and the First, Random, and REPLUG-style likelihood baselines.

Figs.~\ref{fig:lora_alignment} and~\ref{fig:preference_mining} illustrate the two central components of our proposed alignment procedure. Specifically, Fig.~\ref{fig:lora_alignment} depicts the iterative mine–train–re-mine loop, whereas Fig.~\ref{fig:preference_mining} shows how generator feedback is used to construct a positive–negative preference pair for an individual query. In essence, our proposed procedure is enabled by three key mechanisms: generator-guided preference mining uses downstream answer correctness as the supervision signal; batching, early exit, and persistent caching reduce the cost of repeated VLM evaluation; and periodic re-mining refreshes the preference pairs as the reranker evolves.

Fig.~\ref{fig:lora_alignment} shows the iterative alignment loop. At the start, the reranker $E_\theta$ is used to score the Stage~1 candidates, and a preference dataset $\mathcal{D}_1$ is mined from those scores. The reranker then undergoes LoRA fine-tuning using $\mathcal{D}_1$ for $N$ consecutive epochs. After $N$ epochs, the now-improved reranker is used to re-score the same candidates and produce a refreshed dataset $\mathcal{D}_2$. Fine-tuning continues from $\mathcal{D}_2$ for another $N$ epochs, and this mine-then-train cycle repeats until the total number of training epochs $E$ is reached. The key intuition is that as the reranker changes, re-mining can refresh the preference pairs so that training continues to reflect the model's current ranking behavior.

Further, Fig.~\ref{fig:preference_mining} shows how a single preference pair $(d^+, d^-)$ is constructed for one query. The reranker scores all retrieved candidate documents and evaluates them in descending score order using the frozen VLM $\mathcal{G}$. For each candidate, $\mathcal{G}$ generates an answer conditioned on the image, the question, and that candidate as context. The first candidate for which $\mathcal{G}$ produces the correct answer is taken as $d^+$, and the first candidate for which $\mathcal{G}$ produces a wrong answer is taken as $d^-$. Once both documents have been identified, the evaluation terminates through an early-exit mechanism, thereby avoiding unnecessary VLM calls for the remaining candidates. If the entire candidate pool is examined without finding both a positive and a negative document (e.g., when all candidates produce the same correctness outcome), the query is skipped and contributes no preference pair for that mining round.

Subsection~\ref{sec:alg_loop} presents the iterative loop. Section~\ref{sec:alg_loss} defines the three alignment loss functions. Subsection~\ref{sec:alg_mine} details the preference mining subroutine for all four strategies. 

\vspace{-3mm}
\subsection{Iterative Reranker Alignment}
\label{sec:alg_loop}

\begin{algorithm}[t]
\caption{Iterative LoRA Fine-Tuning with Periodic Re-mining}
\label{alg:main_loop}
\begin{algorithmic}[1]
\Require Queries $\mathcal{Q}$, base cross-encoder $E_{\theta_\mathrm{base}}$, frozen VLM $\mathcal{G}$
\Require Pool cap $K'$, re-mine frequency $N$, epochs $E$, learning rate $\eta$, method
\State randomly initialize $\theta_\mathrm{LoRA}$; freeze $\theta_\mathrm{base}$ \label{line:main:l1}
\State $\mathcal{D} \leftarrow \mathrm{PreferenceMine}(\mathcal{Q},\, E_{\theta_\mathrm{base}},\, \mathcal{G},\, K')$
\State $E_\theta \leftarrow E_{\theta_\mathrm{base} + \theta_\mathrm{LoRA}}$ \label{line:main:l3}
\For{$e = 1$ \textbf{to} $E$} \label{line:main:l4}
    \If{$N > 0$ \textbf{and} $(e-1) \bmod N = 0$ \textbf{and} $e > 1$}
        \State $\mathcal{D} \leftarrow \mathrm{PreferenceMine}(\mathcal{Q},\, E_\theta,\, \mathcal{G},\, K')$
    \EndIf
    \For{each $(q_\mathrm{text}, d^+, d^-) \in \mathcal{D}$} 
        \State $s^+ \leftarrow s_\theta(q_\mathrm{text},\, d^+)$
        \State $s^- \leftarrow s_\theta(q_\mathrm{text},\, d^-)$
        \State $\mathcal{L} \leftarrow \mathrm{ComputeLoss}(s^+,\, s^-,\, \mathrm{method})$
        \State update $\theta_\mathrm{LoRA}$ via $\mathrm{AdamW}$ on $\nabla_{\theta_\mathrm{LoRA}}\mathcal{L}$ with learning rate $\eta$ \label{line:main:l12}
    \EndFor
\EndFor
\State \Return final model $E_\theta$
\end{algorithmic}
\end{algorithm}

Algorithm 1 describes the reranker-alignment procedure in two phases: initialization and iterative training. In the \emph{initialization phase} (Lines~\ref{line:main:l1}--\ref{line:main:l3}), the LoRA parameters $\theta_\mathrm{LoRA}$ are randomly initialized and the preference dataset $\mathcal{D}$ is mined using the unmodified base cross-encoder $E_{\theta_\mathrm{base}}$. In other words, the randomly initialized adapters are not used to rank candidates during the initial mining step, preventing them from introducing arbitrary ordering effects into the first set of preference pairs. The LoRA adapters are then incorporated into the reranker, and the base parameters $\theta_\mathrm{base}$ are frozen. Only $\theta_\mathrm{LoRA}$ is updated during training.

In the \emph{iterative training phase} (Lines~\ref{line:main:l4}--\ref{line:main:l12}), the outer loop over epochs $e$ triggers a re-mining step at the start of epoch $e$ whenever $(e-1) \bmod N = 0$ and $e > 1$. Re-mining replaces stale preference pairs (scored under an older model snapshot) with fresh pairs scored by the current $E_\theta$, so that $d^+$ and $d^-$ reflect the model's present ranking. This can provide a useful refresh when the reranker has changed substantially, but additional refreshes may yield diminishing returns after the model saturates. Within each epoch, the training set $\mathcal{D}$ is iterated over sample by sample (in simulations, in mini-batches of 16). Each training sample is a triple $(q_\mathrm{text},\, d^+,\, d^-)$ for which the cross-encoder produces two scalar logits $s^+ = s_\theta(q_\mathrm{text}, d^+)$ and $s^- = s_\theta(q_\mathrm{text}, d^-)$, which are passed to $\mathrm{ComputeLoss}$ to compute the alignment loss (the specific loss depends on the chosen method: Contrastive Triplet, SFT, or Pairwise DPO, as defined in Section~\ref{sec:alg_loss}). The resulting gradient is then back-propagated through $\theta_\mathrm{LoRA}$, and the parameters are updated using AdamW. After the final epoch, the aligned reranker $E_\theta$ is saved.

\vspace{-3mm}
\subsection{Alignment Loss Functions}
\label{sec:alg_loss}

Once a preference pair $(d^+, d^-)$ has been mined, the reranker is trained to assign a higher score to the document that helps the frozen generator answer correctly than to the document that does not. For each pair, the cross-encoder produces two scalar logits, $s^+ = s_\theta(q_\mathrm{text}, d^+)$ and $s^- = s_\theta(q_\mathrm{text}, d^-)$.

\begin{itemize}
    \item \textbf{SFT:}
    The first objective treats answer utility as a supervised regression target. The positive document is assigned the target score $1$, while the negative document is assigned the target score $0$:
    \begin{equation}
    \mathcal{L}_{\text{SFT}} = \left( s_\theta(q_\mathrm{text}, d^+) - 1 \right)^2 + \left( s_\theta(q_\mathrm{text}, d^-) - 0 \right)^2.
    \end{equation}
    This objective directly pushes useful documents toward a high absolute score and unhelpful documents toward a low absolute score. However, because reranking ultimately depends on relative ordering rather than calibrated score values, these fixed targets impose a stronger constraint than is strictly necessary.
    \item \textbf{Contrastive Triplet Loss:}
    The second objective focuses directly on relative preference. Rather than assigning absolute score targets, the triplet loss trains the reranker by enforcing a relative ordering constraint: the relevance score of the positive document $s^+$ must exceed that of the negative document $s^-$ by at least a margin $m$:
    \begin{equation}
    \mathcal{L}_{\text{triplet}} = \max\!\bigl(0,\; m - (s^+ - s^-)\bigr), \quad m = 0.2.
    \end{equation}
    The $\max(0, \cdot)$ hinge structure implies that the loss is zero whenever $s^+ - s^- \geq m$, i.e., the reranker already ranks $d^+$ sufficiently above $d^-$. A gradient is produced only when this margin is violated, pushing $s^+$ up and $s^-$ down until the gap is restored. This makes triplet loss a natural objective for reranking because it optimizes the positive-over-negative ordering directly without requiring the logits to take any particular absolute values. Further, the margin $m = 0.2$ discourages near-ties by encouraging a small score buffer between preferred and rejected documents.

    \item \textbf{Pairwise DPO:}
    The third objective adapts DPO~\cite{rafailov2023direct} to reranking. Following the score definition in (\ref{eq:score}), the trainable logit is mapped to a bounded DPO-style log-score, $\ell_\theta(q_\mathrm{text},d)=\log\sigma(s_\theta(q_\mathrm{text},d))=\log\sigma(E_\theta([q_\mathrm{text};d]))$, where $\sigma(\cdot)$ denotes the logistic sigmoid. For the frozen reference reranker $E_\mathrm{ref}$, the analogous logit is $s_\mathrm{ref}(q_\mathrm{text},d)=E_\mathrm{ref}([q_\mathrm{text};d])$, yielding $\ell_\mathrm{ref}(q_\mathrm{text},d)=\log\sigma(s_\mathrm{ref}(q_\mathrm{text},d))$. The preference margins are then defined as
    \begin{equation}
    \begin{aligned}
        \Delta_\theta
        &= \ell_\theta(q_\mathrm{text},d^+) - \ell_\theta(q_\mathrm{text},d^-),\\
        \Delta_\mathrm{ref}
        &= \ell_\mathrm{ref}(q_\mathrm{text},d^+) - \ell_\mathrm{ref}(q_\mathrm{text},d^-),
    \end{aligned}
    \end{equation}
    The DPO loss is then given by
    \begin{equation}
        \mathcal{L}_{\mathrm{DPO}}
        =
        -\log \sigma\!\left(
        \beta \bigl[(\Delta_\theta) - (\Delta_\mathrm{ref})\bigr]
        \right).
    \end{equation}
    This objective encourages the LoRA-adapted reranker to prefer $d^+$ over $d^-$ more strongly than the original base reranker does. At the same time, the reference margin anchors the update to the pre-trained model, reducing unnecessary deviation from its initial ranking behavior. The parameter $\beta$ controls the strength of this reference-relative preference update.
\end{itemize}

Together, the above objectives provide three complementary views of alignment: SFT learns absolute utility targets, triplet loss enforces relative ordering, and DPO improves the preferred-over-rejected margin relative to the frozen base reranker. Using the same mined preference pairs across all three objectives allows the experiments to isolate the effect of the supervision signal from the effect of the loss function.

\vspace{-3mm}
\subsection{Preference-Mining Strategies}
\label{sec:alg_mine}

The previous subsection defined how a mined pair $(d^+,d^-)$ is used to optimize the reranker under SFT, triplet, or DPO losses. We now describe how these positive-negative pairs are constructed. This distinction is important because all training objectives operate on the same pairwise format, while the supervision strategy determines what qualifies a document as preferred or rejected. For mining, $\mathcal{C}_q$ is the Stage~1 top-$K$ pool retrieved in (\ref{eq:faiss_search}), and $\mathcal{C}_q^{K'}$ denotes its top-$K'$ subset after sorting by the current reranker score, with $K' \le K$. Algorithm~\ref{alg:data_mining} presents the proposed generator-guided strategy. Algorithm~\ref{alg:baseline_mining} covers the three baselines: \textbf{first}, \textbf{random}, and \textbf{REPLUG-style likelihood (REPLUG-L)}.

\begin{algorithm}[t]
\caption{Generator-Guided Preference Mining (Proposed)}
\label{alg:data_mining}
\begin{algorithmic}[1]
\Require Queries $\mathcal{Q}$, reranker $E_\theta$, frozen VLM $\mathcal{G}$, cap $K'$, batch size $B_m$, cache $\Psi$
\State $\mathcal{D} \leftarrow \emptyset$
\For{each $(q, v, a_\mathrm{gold}) \in \mathcal{Q}$}
    \State $\mathcal{C}_q^{K'} \leftarrow \mathrm{top}\text{-}K'(\mathcal{C}_q; s_\theta(q_\mathrm{text},\cdot))$ \Comment{descending score} \label{line:mine:sort}
    \State $d^+ \leftarrow \emptyset$; $d^- \leftarrow \emptyset$
    \For{$i = 0,\, B_m,\, 2B_m, \ldots$} \label{line:mine:batchloop}
        \State $\mathrm{batch} \leftarrow \mathcal{C}_q^{K'}[i : i + B_m]$
        \For{each $d \in \mathrm{batch}$}
            \State $\mathrm{key}(d) \leftarrow (\mathrm{doc\text{-}id}(d),\, \mathrm{pos}(d))$ \label{line:mine:key}
            \If{$\mathrm{key}(d) \notin \Psi$} \label{line:mine:cachecheck}
                \State $\Psi[\mathrm{key}(d)] \leftarrow \mathcal{G}(v, q, d)$ \label{line:mine:vlm}
            \EndIf
            \If{$\Psi[\mathrm{key}(d)] = a_\mathrm{gold}$ \textbf{and} $d^+ = \emptyset$} \label{line:mine:poscheck}
                \State $d^+ \leftarrow d$
            \EndIf \label{line:mine:posend}
            \If{$\Psi[\mathrm{key}(d)] \neq a_\mathrm{gold}$ \textbf{and} $d^- = \emptyset$} \label{line:mine:negcheck}
                \State $d^- \leftarrow d$
            \EndIf \label{line:mine:negend}
        \EndFor \label{line:mine:innend}
        \If{$d^+ \neq \emptyset$ \textbf{and} $d^- \neq \emptyset$} \label{line:mine:earlyexit}
            \State \textbf{break} \Comment{Early exit: both examples identified}
        \EndIf \label{line:mine:earlyend}
    \EndFor \label{line:mine:batchend}
    \If{$d^+ \neq \emptyset$ \textbf{and} $d^- \neq \emptyset$} \label{line:mine:addcheck}
        \State $\mathcal{D} \leftarrow \mathcal{D} \cup \{(q_\mathrm{text},\, d^+,\, d^-)\}$ \label{line:mine:addpair}
    \EndIf \label{line:mine:addend}
\EndFor
\State \Return $\mathcal{D}$
\end{algorithmic}
\end{algorithm}

\textit{1) Generator-Guided Preference Mining:} Algorithm~\ref{alg:data_mining} specifies the proposed generator-guided mining strategy. Its purpose is not to estimate generic document relevance, but to ask whether each candidate changes the deployed generator's answer behavior. For each query triple $(q, v, a_\mathrm{gold})$, the retained candidate set $\mathcal{C}_q^{K'}$ is first sorted by $s_\theta(q_\mathrm{text},\cdot)$ in descending order (Line~\ref{line:mine:sort}), so that the most confidently ranked documents are evaluated first. The algorithm then iterates over mini-batches of $B_m$ candidates in that order (Lines~\ref{line:mine:batchloop}--\ref{line:mine:batchend}). For each candidate $d$, a cache key is formed from its document identifier and position in the retrieved pool (Line~\ref{line:mine:key}); if the key is absent from the persistent cache $\Psi$, the frozen VLM $\mathcal{G}$ is queried with $(v,q,d)$ and the predicted answer is stored (Lines~\ref{line:mine:cachecheck}--\ref{line:mine:vlm}). If the key is already cached, the stored prediction is reused instead of prompting the VLM again. If the prediction matches $a_\mathrm{gold}$ and no positive has been assigned yet, $d$ is taken as $d^+$ (Lines~\ref{line:mine:poscheck}--\ref{line:mine:posend}). If it does not match and no negative has been assigned yet, $d$ is taken as $d^-$ (Lines~\ref{line:mine:negcheck}--\ref{line:mine:negend}). Once both $d^+$ and $d^-$ are identified, an \emph{early exit} terminates the remaining mini-batches immediately (Lines~\ref{line:mine:earlyexit}--\ref{line:mine:earlyend}), avoiding superfluous VLM calls. A pair is added to $\mathcal{D}$ only when both $d^+$ and $d^-$ are found (Lines~\ref{line:mine:addcheck}--\ref{line:mine:addend}); queries for which the pool contains no correct or no incorrect prediction are silently skipped.

\textit{2) Controlled Baselines:} Algorithm~\ref{alg:baseline_mining} defines the three controlled baselines. All baselines use the same Stage 1 candidate pools, reranker architecture, LoRA training procedure, and alignment losses as the proposed method. The only difference is how $d^+$ and $d^-$ are defined. By comparing the proposed method with these baselines, we can assess whether answer-utility supervision is the key source of the gains. 

The \textbf{random baseline} (Lines~\ref{line:base:random}--\ref{line:base:randomend}) draws $d^+$ and $d^-$ uniformly at random from $\mathcal{C}_q^{K'}$ without replacement. It uses neither reranker scores nor VLM feedback. This baseline therefore measures how the alignment losses behave when the preference labels contain no meaningful supervision and provides a noise-sensitivity reference. The \textbf{first baseline} (Lines~\ref{line:base:first}--\ref{line:base:firstend}) relies only on the current reranker ordering. It selects the highest-scored candidate as $d^+$ and the bottom-ranked candidate as $d^-$, without consulting the VLM. This strategy tests whether simply reinforcing the reranker’s existing semantic ranking is sufficient to improve performance. Because it never prompts the VLM, the selected positive may be semantically relevant without helping the generator answer correctly.

The \textbf{REPLUG-style likelihood baseline (REPLUG-L)} (Lines~\ref{line:base:replug}--\ref{line:base:replugend}) uses the VLM differently from our generator strategy. Rather than asking the VLM to produce the constrained answer output, it scores every candidate with a REPLUG-style LM-likelihood signal: the log-likelihood of $a_\mathrm{gold}$ conditioned on $(v, q, d)$, where $a_\mathrm{gold}$ is provided token-by-token as input rather than being generated:
\begin{equation}
    \ell(d) = \frac{1}{|a_\mathrm{gold}|} \sum_{t=1}^{|a_\mathrm{gold}|} \log p_\mathcal{G}\!\bigl(a_{\mathrm{gold},t} \mid v,\, q,\, d,\, a_{\mathrm{gold},<t}\bigr).
    \label{eq:replug_likelihood}
\end{equation}
Here, $p_\mathcal{G}$ denotes the frozen VLM's next-token probability, $|a_\mathrm{gold}|$ is the number of target answer tokens, $a_{\mathrm{gold},t}$ is the $t$-th token, and $a_{\mathrm{gold},<t}$ denotes preceding target tokens. At each step $t$, the VLM $\mathcal{G}$ is conditioned on $(v, q, d, a_{\mathrm{gold},<t})$, and the log probability assigned to $a_{\mathrm{gold},t}$ is averaged across the target answer tokens to produce $\ell(d)$. This is a REPLUG-style ranking score rather than the full REPLUG LSR objective, because we use it only to choose $d^+$ and $d^-$ instead of training a retriever with a document-level KL loss. A higher $\ell(d)$ means that $\mathcal{G}$ assigns greater probability to the target answer when conditioned on document $d$ and the preceding ground-truth answer tokens. The highest-scoring document is selected as $d^+$ and the lowest-scoring document as $d^-$ (Lines~\ref{line:base:replugargmax}--\ref{line:base:replugargmin}), so all $K'$ candidates must be evaluated before selection and early exit is not possible. In contrast, our generator-guided strategy asks the VLM to produce the constrained answer output and checks whether it matches the ground-truth. Thus, a document may have a high $\ell(d)$ even if the VLM would fail to produce the correct constrained answer without those ground-truth answer tokens.

\begin{algorithm}[t]
    \caption{Baseline Preference Mining}
    \label{alg:baseline_mining}
    \begin{algorithmic}[1]
    \Require Queries $\mathcal{Q}$, reranker $E_\theta$, cap $K'$, method $m$
    \Require If $m=\mathrm{replug\text{-}l}$: frozen VLM $\mathcal{G}$, batch size $B_m$, cache $\Psi$
    \State $\mathcal{D} \leftarrow \emptyset$
    \For{each $(q, v, a_\mathrm{gold}) \in \mathcal{Q}$}
        \State $\mathcal{C}_q^{K'} \leftarrow \mathrm{top}\text{-}K'(\mathcal{C}_q; s_\theta(q_\mathrm{text},\cdot))$ \Comment{descending score}
        \If{$m = \mathrm{random}$} \label{line:base:random}
            \State $d^+, d^- \sim \mathrm{Uniform}(\mathcal{C}_q^{K'})$ without replacement
        \EndIf \label{line:base:randomend}
        \If{$m = \mathrm{first}$} \label{line:base:first}
            \State $d^+ \leftarrow \arg\max_{d \in \mathcal{C}_q^{K'}} s_\theta(q_\mathrm{text},d)$
            \State $d^- \leftarrow \arg\min_{d \in \mathcal{C}_q^{K'}} s_\theta(q_\mathrm{text},d)$
        \EndIf \label{line:base:firstend}
        \If{$m = \mathrm{replug\text{-}l}$} \label{line:base:replug}
            \For{each $d \in \mathcal{C}_q^{K'}$ in mini-batches of $B_m$}
                \State $\mathrm{key}(d) \leftarrow (\mathrm{doc\text{-}id}(d),\, \mathrm{pos}(d))$
                \If{$\mathrm{key}(d) \notin \Psi$}
                    \State $\Psi[\mathrm{key}(d)] \leftarrow \ell(d)$
                \EndIf
            \EndFor
            \State $d^+ \leftarrow \arg\max_{d \in \mathcal{C}_q^{K'}} \Psi[\mathrm{key}(d)]$ \label{line:base:replugargmax}
            \State $d^- \leftarrow \arg\min_{d \in \mathcal{C}_q^{K'}} \Psi[\mathrm{key}(d)]$ \label{line:base:replugargmin}
        \EndIf \label{line:base:replugend}
        \If{$d^+ \neq \emptyset$ \textbf{and} $d^- \neq \emptyset$}
            \State $\mathcal{D} \leftarrow \mathcal{D} \cup \{(q_\mathrm{text},\, d^+,\, d^-)\}$
        \EndIf
    \EndFor
    \State \Return $\mathcal{D}$
    \end{algorithmic}
    \end{algorithm}

\vspace{-3mm}
\section{Experiments}
We first describe the implementation details, followed by the results and analysis.

\subsection{Implementation Details}

\begin{table}[t]
\centering
\caption{Default Hyperparameter Configuration for VQA-X and A-OKVQA}
\label{tab:hyperparameters}
\renewcommand{\arraystretch}{1.1}
\begin{tabular}{p{0.22\linewidth}p{0.29\linewidth}p{0.38\linewidth}}
\toprule
\textbf{Category} & \textbf{Hyperparameter} & \textbf{Value} \\
\midrule
\textbf{Architecture} & VLM Generator & Qwen3-VL-4B-Instruct, Qwen3.5-2B \\
 & Sentence Transformer & Qwen3-Embedding-8B \\
 & Reranker Backbone & GTE-ModernBERT reranker \\
 & Trainable Module & LoRA-adapted reranker \\
\midrule
\textbf{LoRA Adapter} & Rank ($r$) & 16 \\
 & Alpha ($\alpha$) & 32 \\
 & Dropout & 0.1 \\
\midrule
\textbf{Optimization} & Optimizer & AdamW \\
 & Learning Rate & $5 \times 10^{-5}$ \\
 & Training Epochs & 2 \\
 & Batch Size & 16 \\
\midrule
\textbf{Objective} & DPO Beta ($\beta$) & 0.1 \\
\midrule
\textbf{Mining} & Stage~2 Pool Size & $K'=10$ of 100 candidates \\
 & Mining Batch Size & 16 \\
 & Re-mining Frequency & 0 (disabled) \\
 & REPLUG-L Pool Size & $K'=10$ of 100 candidates \\
\bottomrule
\end{tabular}
\vspace{-3mm}
\end{table}

We apply one default hyperparameter configuration across both datasets, summarized in Table~\ref{tab:hyperparameters}; experiments that deviate from these defaults state the changed value explicitly. Stage~1 retrieves 100 candidates per query using the \texttt{Qwen3-Embedding-8B} dense encoder with a FAISS index. For Stage~2, the cross-encoder backbone is \texttt{Alibaba-NLP/gte-reranker-modernbert-base}. Preference mining uses the top 10 retrieved candidates by default ($K' = 10$), with re-mining disabled unless otherwise stated. We evaluate two frozen VLM generators, \texttt{Qwen3-VL-4B-Instruct} and \texttt{Qwen3.5-2B}, to assess whether the proposed training-signal strategy generalizes across generators with different capacities.

The reranker is fine-tuned with LoRA using rank $r=16$, scaling factor $\alpha=32$, and dropout $0.1$. Optimization uses AdamW with a learning rate of $5\times10^{-5}$, batch size of 16, and 2 epochs unless otherwise stated. For Pairwise DPO, the inverse-temperature parameter is set to $\beta=0.1$.

We do not apply periodic re-mining in the main comparison (\texttt{remine\_every=0}); re-mining frequency is studied separately in Table~\ref{tab:remine_freq}. The REPLUG-style likelihood baseline scores all 10 candidates by ground-truth-token log-likelihood.

Stage~1 and Stage~2 are split-aware: the retrieval index is built from training-split documents only, while Stage~2 loads the selected dataset split to recover images and labels for the candidate question identifiers. A-OKVQA is evaluated as a multiple-choice task: the prompt lists choices as \texttt{A)}, \texttt{B)}, etc. Although the Qwen model cards and generation configs provide sampling-oriented default or recommended values, our evaluation uses deterministic greedy decoding by setting \texttt{do\_sample=False} with \texttt{max\_new\_tokens=10}; temperature and top-$p$ sampling are therefore not used. The output is parsed first as a leading valid answer letter, with a fallback regex over standalone letters. Accuracy is exact match against \texttt{correct\_choice\_idx}. VQA-X is evaluated on yes/no questions: the prompt instructs the VLM to answer strictly with ``yes'' or ``no'', generation uses the same deterministic decoding settings, outputs are lowercased, simple markers such as \texttt{*}, \texttt{\_}, and periods are removed before \texttt{yes}/\texttt{no} matching, and accuracy is the standard VQA soft score $\min(\#\text{matching annotators}/3,1)$. During binary preference mining for VQA-X, a prediction is treated as correct when this VQA score reaches 1.

\vspace{-3mm}
\subsection{Results and Analysis}

\begin{table}[t]
    \caption{Impact of Training-Signal Construction Across Alignment Loss Functions (2 Epochs, Generator: Qwen3-VL-4B-Instruct, Candidate Pool: Top~10). Results reported as mean\,$\pm$\,std across 3 random seeds.}
    \label{tab:alignment_comparison}
    \centering
    \small
    \begin{tabular}{llcc}
    \toprule
    \textbf{\shortstack[l]{Alignment\\Loss}} & \textbf{\shortstack[l]{Training\\Signal}} & \raisebox{0.5\normalbaselineskip}{\textbf{VQA-X (\%)}} & \raisebox{0.5\normalbaselineskip}{\textbf{A-OKVQA (\%)}} \\
    \midrule
    Base & No training & 94.75 & 85.04 \\
    \midrule
    \multirow{4}{*}{\shortstack[l]{Contrastive\\(Triplet)}} & First   & 94.72$\pm$0.22 & 85.07$\pm$0.11 \\
     & Random                                                    & 92.96$\pm$0.14 & 82.77$\pm$0.72 \\
     & REPLUG-L                                                  & 95.85$\pm$0.10 & 82.03$\pm$0.28 \\
     & Generator                                        & \textbf{96.27$\pm$0.03} & \textbf{87.00$\pm$0.13} \\
    \midrule
    \multirow{4}{*}{\shortstack[l]{Pairwise\\DPO}} & First       & 94.82$\pm$0.04 & 85.15$\pm$0.05 \\
     & Random                                                    & 93.84$\pm$0.82 & 83.63$\pm$1.16 \\
     & REPLUG-L                                                  & 95.79$\pm$0.09 & 82.40$\pm$0.27 \\
     & Generator                                        & \textbf{96.11$\pm$0.04} & \textbf{87.16$\pm$0.03} \\
    \midrule
    \multirow{4}{*}{SFT} & First                                 & 94.75$\pm$0.05 & 85.16$\pm$0.02 \\
     & Random                                                    & 93.35$\pm$0.37 & 80.95$\pm$0.72 \\
     & REPLUG-L                                                  & 95.69$\pm$0.04 & 82.65$\pm$0.14 \\
     & Generator                                         & \textbf{95.73$\pm$0.01} & \textbf{85.75$\pm$0.08} \\
    \bottomrule
    \end{tabular}
    \end{table}
    
    \begin{table}[t]
    \caption{Impact of Training-Signal Construction Across Alignment Loss Functions (2 Epochs, Generator: Qwen3.5-2B, Candidate Pool: Top~10). Results reported as mean\,$\pm$\,std across 3 random seeds.}
    \label{tab:alignment_comparison_2b}
    \centering
    \small
    \begin{tabular}{llcc}
    \toprule
    \textbf{\shortstack[l]{Alignment\\Loss}} & \textbf{\shortstack[l]{Training\\Signal}} & \raisebox{0.5\normalbaselineskip}{\textbf{VQA-X (\%)}} & \raisebox{0.5\normalbaselineskip}{\textbf{A-OKVQA (\%)}} \\
    \midrule
    Base & No training & 90.28 & 79.83 \\
    \midrule
    \multirow{4}{*}{\shortstack[l]{Contrastive\\(Triplet)}} & First   & 90.24$\pm$0.13 & 79.76$\pm$0.11 \\
     & Random                                                    & 89.13$\pm$2.01 & 77.02$\pm$1.37 \\
     & REPLUG-L                                                  & 92.44$\pm$0.30 & 76.98$\pm$0.07 \\
     & Generator                                                 & \textbf{94.82$\pm$0.11} & \textbf{84.67$\pm$0.21} \\
    \midrule
    \multirow{4}{*}{\shortstack[l]{Pairwise\\DPO}} & First       & 90.20$\pm$0.18 & 79.99$\pm$0.08 \\
     & Random                                                    & 89.49$\pm$0.78 & 79.50$\pm$0.61 \\
     & REPLUG-L                                                  & 92.76$\pm$0.18 & 76.97$\pm$0.11 \\
     & Generator                                                 & \textbf{93.32$\pm$0.04} & \textbf{82.50$\pm$0.06} \\
    \midrule
    \multirow{4}{*}{SFT} & First                                 & 90.12$\pm$0.16 & 79.94$\pm$0.08 \\
     & Random                                                    & 89.27$\pm$0.71 & 76.72$\pm$0.63 \\
     & REPLUG-L                                                    & 92.13$\pm$0.14 & 76.95$\pm$0.12 \\
     & Generator                                                 & \textbf{92.54$\pm$0.10} & \textbf{80.75$\pm$0.16} \\
    \bottomrule
    \end{tabular}
    \vspace{-4mm}
    \end{table}

We evaluate our proposed framework across three dimensions: (i) the superiority of the generator-guided training signal over alternative mining strategies under all three alignment losses (Tables~\ref{tab:alignment_comparison} and~\ref{tab:alignment_comparison_2b}), (ii) the robustness of the generator strategy to the size of the Stage~1 candidate pool (Tables~\ref{tab:triplet_pool_sensitivity} and~\ref{tab:triplet_pool_sensitivity_2b}), and (iii) the impact of periodic preference re-mining on accuracy (Table~\ref{tab:remine_freq}).

We note that direct comparison with existing retriever–generator alignment systems is difficult because prior methods often target text-only tasks, use different retrieval corpora and generators, or require joint retriever–generator training or an additional trainable bridge module. We therefore adopt a controlled comparison in which the Stage 1 candidate pool, reranker architecture, frozen generator, training configuration, and alignment loss are held fixed, while only the strategy used to construct $(d^+,d^-)$ is varied. Under this design, the First baseline tests whether reinforcing the reranker’s existing score order is sufficient; the Random baseline measures the behavior of pairwise training under uninformative preference labels; and the REPLUG-style likelihood baseline represents likelihood-based generator-aware supervision adapted from prior work \cite{shi2024replug}. Our proposed method differs by assigning preferences according to whether each candidate enables the frozen VLM to produce the correct constrained answer output.  

All trained results are reported as mean $\pm$ standard deviation over three random seeds. The seeds vary the LoRA initialization, training-data order, and random pair sampling where applicable, while the frozen generators and deterministic decoding procedure remain unchanged. We therefore assess a strategy based not only on its mean accuracy, but also on whether its gains are consistent across datasets, generators, alignment losses, and random seeds. The following subsections analyze the three dimensions in sequence.

\subsubsection{Generator-Guided Mining Outperforms All Baselines}

Tables~\ref{tab:alignment_comparison} and~\ref{tab:alignment_comparison_2b} compare all training-signal strategies under three alignment losses for Qwen3-VL-4B-Instruct and Qwen3.5-2B, respectively, on VQA-X and A-OKVQA. The base cross-encoder selects the first document as the context and achieves 94.75\% / 85.04\% (Qwen3-VL-4B-Instruct) and 90.28\% / 79.83\% (Qwen3.5-2B), establishing the no-training reference that an effective strategy must surpass.

Our proposed generator strategy achieves the highest accuracy in the evaluated cells of both tables, regardless of alignment loss, generator, or dataset. With Qwen3-VL-4B-Instruct, it reaches 96.27\% / 87.00\% under Contrastive (Triplet) and 96.11\% / 87.16\% under Pairwise DPO, with SFT trailing at 95.73\% / 85.75\%. With the smaller Qwen3.5-2B, Contrastive training achieves the best result (94.82\% / 84.67\%), followed by Pairwise DPO (93.32\% / 82.50\%) and SFT (92.54\% / 80.75\%). The consistent advantage across both generators suggests that the improvement stems from the quality of the training signal rather than any capacity-specific effect.

The controlled baselines further clarify which supervision signal matters. The First strategy represents rank-order self-training: it reinforces the reranker's own semantic ordering without answer-level feedback, and it often fails to improve over the untrained base (e.g., 94.72\% / 85.07\% vs.\ the base 94.75\% / 85.04\% for Qwen3-VL-4B-Instruct under Contrastive). The Random strategy tests whether gains arise merely from LoRA fine-tuning with pairwise objectives; its frequent degradation below the base (e.g., 92.96\% on VQA-X for Qwen3-VL-4B-Instruct under Contrastive, and 89.13\% for Qwen3.5-2B) shows that arbitrary pair construction can harm the reranker. The REPLUG-style likelihood baseline (REPLUG-L) is the closest prior-work-connected comparison, adapting likelihood-based generator-aware retrieval supervision. It improves VQA-X over the base across all configurations (e.g., 95.85\%, 95.79\%, and 95.69\% for Qwen3-VL-4B-Instruct under the three losses), but it hurts A-OKVQA accuracy relative to the base in these experiments (e.g., 82.03\%--82.65\% vs.\ the base 85.04\% for Qwen3-VL-4B-Instruct). This asymmetry suggests that ground-truth-token likelihood is an imperfect proxy for constrained-answer correctness: a high likelihood of $a_\mathrm{gold}$ when preceding answer tokens are supplied does not reliably indicate that $\mathcal{G}$ would produce the correct constrained answer.

Across the three alignment losses, Contrastive (Triplet) emerges as a strong default: it is consistently competitive and becomes the clear best-performing objective with Qwen3.5-2B, which motivates using it in the pool-sensitivity and re-mining studies below. At the same time, Pairwise DPO reaches comparable accuracy with Qwen3-VL-4B-Instruct under the same generator-guided mining signal. Thus, the main finding is not that triplet loss universally dominates, but that answer-utility preference mining is the decisive factor, with triplet loss serving as a reliable and simple instantiation.

\subsubsection{Robustness to Stage~1 Candidate Pool Size}

Tables~\ref{tab:triplet_pool_sensitivity} and~\ref{tab:triplet_pool_sensitivity_2b} investigate how sensitive each training signal is to $K'$, the number of Stage~1 candidates used during preference mining, under Contrastive (Triplet) training at 2 epochs for Qwen3-VL-4B-Instruct and Qwen3.5-2B, respectively.

The results show that the generator strategy is stable across pool sizes, though the degree of stability differs by generator. For Qwen3-VL-4B-Instruct, VQA-X accuracy increases modestly from 95.88\% at Top~2 to 96.27\% at Top~10 (a range of 0.39 pp), and A-OKVQA grows from 86.22\% to 87.00\% (0.78 pp). For Qwen3.5-2B, the variation is more pronounced: VQA-X rises from 92.76\% at Top~2 to 94.82\% at Top~10 (2.06 pp), and A-OKVQA from 81.45\% to 84.67\% (3.22 pp), suggesting that Qwen3.5-2B, despite being smaller and from a different model generation, benefits more from a richer candidate pool when constructing preference pairs. Notably, our proposed generator strategy with the smallest pool of Top~2 already outperforms the evaluated baselines at their largest pool of Top~10 across both generators and datasets. This aligns with recent budget-aware RAG findings that the utility of retrieved chunks, not merely the number of chunks used, is central to effective retrieval augmentation \cite{al2026budget}; in our setting, the quality of the training signal matters more than the quantity of candidates evaluated.

Further inspection shows that the Random strategy has the highest variance across pool sizes (e.g., 87.73\%--89.13\% on VQA-X for Qwen3.5-2B), because the quality of the sampled pair depends heavily on which documents happen to fall inside the pool. The First strategy is essentially insensitive to pool size, remaining near base-level performance regardless of $K'$, because it always selects the top-1 and bottom-1 of the ranked list and the pool boundary only marginally affects the bottom candidate. REPLUG-L scores all $K'$ candidates in the pool, so it naturally uses more information as the pool grows; even so, at Top~10 it remains below our proposed generator strategy by 0.42 pp on VQA-X and 4.97 pp on A-OKVQA for Qwen3-VL-4B-Instruct (95.85\% vs.\ 96.27\% and 82.03\% vs.\ 87.00\%). For Qwen3.5-2B the gap is even larger: 2.38 pp on VQA-X (92.44\% vs.\ 94.82\%) and 7.69 pp on A-OKVQA (76.98\% vs.\ 84.67\%). This suggests that REPLUG-L's shortfall is not only a matter of how many candidates it evaluates, but also that its ground-truth-token likelihood signal can be a weaker proxy for downstream answer correctness than the direct binary judgment used by the generator strategy.

\begin{table}[t]
    \caption{Sensitivity to Stage~1 Candidate Pool in Contrastive (Triplet) Training (2 Epochs, Generator: Qwen3-VL-4B-Instruct). Results reported as mean\,$\pm$\,std across 3 random seeds.}
    \label{tab:triplet_pool_sensitivity}
    \centering
    \small
    \begin{tabular}{llcc}
    \toprule
    \textbf{\shortstack[l]{Training\\Signal}} & \textbf{\shortstack[l]{Candidate\\Pool}} & \raisebox{0.5\normalbaselineskip}{\textbf{VQA-X (\%)}} & \raisebox{0.5\normalbaselineskip}{\textbf{A-OKVQA (\%)}} \\
    \midrule
    \multirow{5}{*}{\shortstack[l]{Generator\\(Ours)}} & Top 2  & 95.88$\pm$0.03 & 86.22$\pm$0.09 \\
     & Top 4                              & 96.06$\pm$0.03 & 86.68$\pm$0.06 \\
     & Top 6                              & 96.16$\pm$0.03 & 86.88$\pm$0.05 \\
     & Top 8                              & 96.18$\pm$0.08 & 86.93$\pm$0.06 \\
     & Top 10                             & \textbf{96.27$\pm$0.03} & \textbf{87.00$\pm$0.13} \\
    \midrule
    \multirow{5}{*}{First} & Top 2       & 94.73$\pm$0.04 & 85.05$\pm$0.01 \\
     & Top 4                              & 94.76$\pm$0.09 & 85.05$\pm$0.03 \\
     & Top 6                              & 94.68$\pm$0.07 & 85.03$\pm$0.02 \\
     & Top 8                              & 94.61$\pm$0.18 & 85.06$\pm$0.05 \\
     & Top 10                             & 94.72$\pm$0.22 & 85.07$\pm$0.11 \\
    \midrule
    \multirow{5}{*}{Random} & Top 2      & 92.62$\pm$1.05 & 82.24$\pm$0.60 \\
     & Top 4                              & 93.50$\pm$0.85 & 82.50$\pm$0.48 \\
     & Top 6                              & 93.07$\pm$1.31 & 84.15$\pm$0.25 \\
     & Top 8                              & 93.18$\pm$0.42 & 82.61$\pm$0.92 \\
     & Top 10                             & 92.96$\pm$0.14 & 82.77$\pm$0.72 \\
    \midrule
    \multirow{5}{*}{REPLUG-L} & Top 2    & 95.52$\pm$0.03 & 81.45$\pm$0.20 \\
     & Top 4                              & 95.75$\pm$0.10 & 81.58$\pm$0.38 \\
     & Top 6                              & 95.86$\pm$0.10 & 81.89$\pm$0.11 \\
     & Top 8                              & 95.83$\pm$0.10 & 82.04$\pm$0.09 \\
     & Top 10                             & 95.85$\pm$0.10 & 82.03$\pm$0.28 \\
    \bottomrule
    \end{tabular}
    \end{table}
    
    \begin{table}[t]
    \caption{Sensitivity to Stage~1 Candidate Pool in Contrastive (Triplet) Training (2 Epochs, Generator: Qwen3.5-2B). Results reported as mean\,$\pm$\,std across 3 random seeds.}
    \label{tab:triplet_pool_sensitivity_2b}
    \centering
    \small
    \begin{tabular}{llcc}
    \toprule
    \textbf{\shortstack[l]{Training\\Signal}} & \textbf{\shortstack[l]{Candidate\\Pool}} & \raisebox{0.5\normalbaselineskip}{\textbf{VQA-X (\%)}} & \raisebox{0.5\normalbaselineskip}{\textbf{A-OKVQA (\%)}} \\
    \midrule
    \multirow{5}{*}{\shortstack[l]{Generator\\(Ours)}} & Top 2  & 92.76$\pm$0.02 & 81.45$\pm$0.02 \\
     & Top 4                              & 93.01$\pm$0.05 & 81.87$\pm$0.16 \\
     & Top 6                              & 93.11$\pm$0.06 & 82.02$\pm$0.01 \\
     & Top 8                              & 93.29$\pm$0.04 & 81.97$\pm$0.02 \\
     & Top 10                             & \textbf{94.82$\pm$0.11} & \textbf{84.67$\pm$0.21} \\
    \midrule
    \multirow{5}{*}{First} & Top 2       & 90.28$\pm$0.05 & 79.85$\pm$0.03 \\
     & Top 4                              & 90.25$\pm$0.12 & 79.76$\pm$0.06 \\
     & Top 6                              & 90.24$\pm$0.05 & 79.85$\pm$0.01 \\
     & Top 8                              & 90.33$\pm$0.24 & 79.74$\pm$0.11 \\
     & Top 10                             & 90.24$\pm$0.13 & 79.76$\pm$0.11 \\
    \midrule
    \multirow{5}{*}{Random} & Top 2      & 87.73$\pm$0.85 & 78.18$\pm$0.69 \\
     & Top 4                              & 88.94$\pm$1.64 & 76.64$\pm$1.12 \\
     & Top 6                              & 88.34$\pm$1.00 & 77.76$\pm$0.90 \\
     & Top 8                              & 88.80$\pm$1.77 & 78.30$\pm$1.12 \\
     & Top 10                             & 89.13$\pm$2.01 & 77.02$\pm$1.37 \\
    \midrule
    \multirow{5}{*}{REPLUG-L} & Top 2    & 91.58$\pm$0.27 & 76.89$\pm$0.22 \\
     & Top 4                              & 91.92$\pm$0.23 & 76.91$\pm$0.25 \\
     & Top 6                              & 92.17$\pm$0.12 & 77.13$\pm$0.06 \\
     & Top 8                              & 92.33$\pm$0.20 & 77.10$\pm$0.07 \\
     & Top 10                             & 92.44$\pm$0.30 & 76.98$\pm$0.07 \\
    \bottomrule
    \end{tabular}
    \vspace{-3mm}
    \end{table}

    \begin{table}[t]
        \caption{Impact of Re-mining Frequency on Downstream Accuracy (Candidate Pool: Top~10). Results reported as mean\,$\pm$\,std across 3 random seeds.}
        \label{tab:remine_freq}
        \centering
        \small
        \begin{tabular}{llcc}
        \toprule
        \raisebox{0.5\normalbaselineskip}{\textbf{Generator}} & \textbf{\shortstack[l]{\# of\\Re-mines}} & \raisebox{0.5\normalbaselineskip}{\textbf{VQA-X (\%)}} & \raisebox{0.5\normalbaselineskip}{\textbf{A-OKVQA (\%)}} \\
        \midrule
        \multirow{3}{*}{\shortstack[l]{Qwen3-VL-4B\\-Instruct}}
        & 0  & 96.27$\pm$0.03 & 87.00$\pm$0.13 \\
        & 2  & \textbf{97.45$\pm$0.09} & 89.01$\pm$0.23 \\
        & 4  & \textbf{97.44$\pm$0.09} & \textbf{89.58$\pm$0.04} \\
        \midrule
        \multirow{3}{*}{Qwen3.5-2B}
        & 0  & 94.43$\pm$0.09 & 82.24$\pm$0.14 \\
        & 2  & 94.82$\pm$0.11 & 84.47$\pm$0.30 \\
        & 4  & \textbf{95.16$\pm$0.13} & \textbf{84.67$\pm$0.21} \\
        \bottomrule
        \end{tabular}
        \vspace{-3mm}
        \end{table}

\subsubsection{Impact of Periodic Re-mining}

Table~\ref{tab:remine_freq} examines the effect of re-mining frequency on downstream accuracy over 4 training epochs using the generator strategy and Contrastive (Triplet) loss. Three configurations are compared: no re-mining (the preference dataset is fixed throughout training), re-mining twice (once at the start before training and once at mid-training), and re-mining four times (once per epoch).

Periodic re-mining generally enhances the performance compared with using a fixed preference dataset, but the benefit of increasing the re-mining frequency is model- and dataset-dependent. For Qwen3-VL-4B-Instruct on VQA-X, re-mining twice and four times are effectively tied within the reported seed variation (97.45$\pm$0.09\% vs.\ 97.44$\pm$0.09\%), indicating saturation rather than a meaningful gain from additional refreshes. In contrast, re-mining four times yields the best A-OKVQA accuracy for Qwen3-VL-4B-Instruct (89.58\% vs.\ 87.00\% without re-mining, +2.58 pp). For Qwen3.5-2B, re-mining four times produces the best results on both datasets (95.16\% on VQA-X, +0.73 pp; 84.67\% on A-OKVQA, +2.43 pp over no re-mining). These results suggest that refreshing stale preference pairs can improve training, while the marginal gains from more frequent refreshes can saturate once the reranker stabilizes. This is also evident from the fact that the incremental gain from four rather than two re-mines is small and dataset-dependent. For Qwen3-VL-4B-Instruct, VQA-X is effectively tied (97.44\% vs.\ 97.45\%), while four re-mines add 0.57 pp on A-OKVQA (89.58\% vs.\ 89.01\%). For Qwen3.5-2B, four re-mines give modest gains over two on both datasets (+0.34 pp VQA-X, +0.20 pp A-OKVQA). Re-mining also adds cost because each refresh must re-rank the mining queries, rebuild pairs from the current top-$K'$ subset, and evaluate any newly reached uncached query-document pairs. Thus, re-mining twice offers a favorable accuracy-cost balance, while additional refreshes should be treated as optional.

\section{Limitations}

This work presents several opportunities for further development. First, although the framework eliminates the need for human document-level relevance annotations, preference mining still relies on answer-labeled training examples to determine whether a candidate document helps the frozen generator produce the correct answer. Future work could reduce this dependence by using self-supervised consistency signals, confidence-based pseudo-labels, or other forms of weak supervision.

Second, preference mining introduces additional computational cost because the VLM must evaluate multiple candidate documents for each query, and periodic re-mining repeats this process as the reranker evolves. This overhead can be further reduced through more selective candidate evaluation, adaptive re-mining schedules, and lightweight surrogate models that approximate the generator’s utility judgments.

Third, the learned reranker is intentionally generator-specific, as it is optimized to select evidence that improves the frozen VLM's behavior during training rather than to learn a universal notion of document relevance. Future work could improve transferability by incorporating preferences from multiple generators or by developing generator-agnostic utility representations that require only limited adaptation to a new VLM.

Finally, the current pairwise training procedure uses only queries whose candidate pools contain at least one helpful and one unhelpful document. While this requirement produces clean positive–negative pairs, it excludes queries for which all candidates yield the same outcome and uses only a limited portion of the available utility information. This limitation could be addressed through graded, listwise, or multi-document utility supervision that uses more of the information available in the candidate pool.

\section{Conclusion}

This paper introduced a \emph{generator-in-the-loop alignment framework} for multimodal RAG. Our proposed two-stage framework separates modality bridging from answer-utility alignment: Stage~1 uses a frozen VLM to generate a HyDE-style text query for dense recall, while Stage~2 trains a LoRA-adapted cross-encoder reranker from answer-supervised preference pairs mined with the same frozen generator. We demonstrated that across VQA-X and A-OKVQA datasets, generator-guided preferences improve over rank-order, random, and REPLUG-style likelihood baselines under multiple alignment losses, suggesting that downstream answer behavior is a useful signal for reranker alignment.
The results also showed that periodic re-mining can improve accuracy by refreshing preference pairs as the reranker changes, although the marginal value of more frequent refreshes depends on the model and dataset. Overall, the findings support a generator-centered view of RAG reranking: the most useful evidence is not the passage semantically closest to the query, but the passage that helps the deployed generator answer correctly. 

\section{Acknowledgments}

Generative AI tools, including ChatGPT (OpenAI) \cite{openai_chatgpt} and Claude (Anthropic) \cite{anthropic_claude}, were used to assist with the preparation and refinement of manuscript text, including sentence-level language improvement. All technical content was developed by the authors, and the full manuscript was carefully reviewed and verified by the authors to ensure accuracy and correctness.

\bibliographystyle{IEEEtran}
\bibliography{tai}

\end{document}